%% file: main.tex
\documentclass[lettersize,journal]{IEEEtran}

\title{Transient Stability Analysis with Physics-Informed Neural Networks}

\author{Jochen~Stiasny,~\IEEEmembership{Student~Member,~IEEE},
        Georgios~S.~Misyris,~\IEEEmembership{Student~Member,~IEEE},        Spyros~Chatzivasileiadis,~\IEEEmembership{Senior Member,~IEEE}%
\thanks{J. Stiasny, G. S. Misyris and S. Chatzivasileiadis are with the Technical University of Denmark, Department of Electrical Engineering, Kgs. Lyngby, Denmark (emails: \{jbest, gmisy, spchatz\}@elektro.dtu.dk).}%
\thanks{This work is supported by Innovation Fund Denmark through the multiDC project (grant no. \mbox{6154-00020B}) and by the ERC Starting Grant VeriPhIED, Grant Agreement No. 949899.}}

\usepackage{cite}
\usepackage{amsmath}
\usepackage{algorithmic}
\usepackage{array}
\ifCLASSOPTIONcompsoc
 \usepackage[caption=false,font=normalsize,labelfont=sf,textfont=sf]{subfig}
\else
 \usepackage[caption=false,font=footnotesize]{subfig}
\fi
\usepackage{url}
\usepackage[utf8]{inputenc}
\usepackage[english]{babel}
\usepackage{xcolor}
\usepackage{tikz}
\usetikzlibrary{shapes,arrows}
\usetikzlibrary{calc}
\usepackage[nooldvoltagedirection]{circuitikz}
\usepackage{mathdots}
\usepackage{tabularx}
\usepackage{booktabs}       
\usepackage{siunitx}

\usepackage[capitalise]{cleveref}
\usepackage{bm}

\usepackage{tikz,pgf} 
\usepackage{lipsum}
\usetikzlibrary{fit}
\usepackage{pgfplotstable}
\pgfplotsset{compat = 1.15}
\usepgfplotslibrary{statistics}
\usetikzlibrary{intersections}
\usepgfplotslibrary{fillbetween}
\usepgfplotslibrary{colorbrewer}
\usetikzlibrary{pgfplots.statistics, pgfplots.colorbrewer} 
\usepgfplotslibrary{statistics}
\pgfplotsset{cycle list/OrRd-3}
\usepgfplotslibrary{groupplots}
\usetikzlibrary{plotmarks}
\usetikzlibrary{decorations.pathreplacing}
\usetikzlibrary{spy}

\DeclareSIUnit\pu{p.u.}

\definecolor{color_bus1}{rgb}{0.12156862745098039, 0.47058823529411764, 0.7058823529411765}
\definecolor{color_bus2}{rgb}{0.2, 0.6274509803921569, 0.17254901960784313}
\definecolor{color_bus3}{rgb}{0.8901960784313725, 0.10196078431372549, 0.10980392156862745}
\definecolor{color_bus4}{rgb}{1.0, 0.4980392156862745, 0.0}
\definecolor{color_bus7}{rgb}{0.41568627450980394, 0.23921568627450981, 0.6039215686274509}
\definecolor{color_bus9}{rgb}{0.6941176470588235, 0.34901960784313724, 0.1568627450980392}

\definecolor{color_NN}{rgb}{0.992, 0.906, 0.78125}
\definecolor{color_dtNN}{rgb}{0.988, 0.730, 0.516}
\definecolor{color_PINN}{rgb}{0.887, 0.289, 0.199}

\newcommand{\xtrue}{\ensuremath{\bm{x}}}

\newcommand{\xtrueij}{\ensuremath{x^i_j}}
\newcommand{\xtruedt}{\ensuremath{\frac{d}{dt}\bm{x}}}
\newcommand{\xhat}{\ensuremath{\hat{\bm{x}}}}

\newcommand{\xhatij}{\ensuremath{\hat{x}^i_j}}
\newcommand{\xhatdt}{\ensuremath{\frac{d}{dt}\hat{\bm{x}}}}

\newcommand{\xhatdtij}{\ensuremath{\frac{d}{dt}\hat{x}^i_j}}
\newcommand{\xinitial}{\ensuremath{\bm{x}_0}}
\newcommand{\inputVariable}{\ensuremath{\bm{u}}}

\newcommand{\vectorf}{\ensuremath{\bm{f}}}
\newcommand{\vectorg}{\ensuremath{\bm{g}}}
\newcommand{\vectorfi}{\ensuremath{f^i}}

\newcommand{\neuronsPerLayer}{\ensuremath{N_L}}
\newcommand{\numberLayers}{\ensuremath{N_{K}}}
\newcommand{\numberDatapoints}{\ensuremath{N_{x}}}
\newcommand{\numberCollocationpoints}{\ensuremath{N_{f}}}

\newcommand{\activationFunction}{\ensuremath{\phi}}

\newcommand{\lossx}{\ensuremath{\mathcal{L}_{x}}}
\newcommand{\lossxi}{\ensuremath{\mathcal{L}_{x}^i}}
\newcommand{\lossdti}{\ensuremath{\mathcal{L}_{dt}^i}}
\newcommand{\lossf}{\ensuremath{\mathcal{L}_{f}}}
\newcommand{\lossfi}{\ensuremath{\mathcal{L}_{f}^i}}
\newcommand{\lossWeightxi}{\ensuremath{\lambda_{x}^i}}
\newcommand{\lossWeightdti}{\ensuremath{\lambda_{dt}^i}}
\newcommand{\lossWeightfi}{\ensuremath{\lambda_{f}^i}}

\newcommand{\suchthat}{\ensuremath{\rm s.t.}}

\newcommand{\weights}{\ensuremath{\bm{W}}}
\newcommand{\biases}{\ensuremath{\bm{b}}}
\newcommand{\perceptron}{\ensuremath{\bm{z}}}

\newcommand{\weightindex}[1]{\ensuremath{\weights_{#1}}}
\newcommand{\biasindex}[1]{\ensuremath{\biases_{#1}}}
\newcommand{\perceptronindex}[1]{\ensuremath{\perceptron_{#1}}}

\begin{document}

\maketitle
\IEEEpeerreviewmaketitle

\input{sections/00_abstract}

\begin{IEEEkeywords}
Neural networks, power system dynamic stability, time-domain analysis, transient analysis.
\end{IEEEkeywords}

\input{sections/01_introduction}
\input{sections/02_methodology}
\input{sections/03_case_study}
\input{sections/04_results}

\input{sections/05_discussion}

\input{sections/06_conclusion}

\bibliographystyle{IEEEtran}
\bibliography{references.bib}
\end{document}

%% file: sections/00_abstract.tex
\begin{abstract}
      We explore the possibility to use physics-informed neural networks to drastically accelerate the solution of ordinary differential-algebraic equations that govern the power system dynamics. When it comes to transient stability assessment, the traditionally applied methods either carry a significant computational burden, require model simplifications, or use overly conservative surrogate models. Conventional neural networks can circumvent these limitations but are faced with high demand of high-quality training datasets, while they ignore the underlying governing equations. Physics-informed neural networks are different: they incorporate the power system differential algebraic equations directly into the neural network training and drastically reduce the need for training data. 
    This paper takes a deep dive into the performance of physics-informed neural networks for power system transient stability assessment. Introducing a new neural network training procedure to facilitate a thorough comparison, we explore how physics-informed neural networks compare with conventional differential-algebraic solvers and classical neural networks in terms of computation time, requirements in data, and prediction accuracy. 
    We illustrate the findings on the Kundur two-area system, and assess the opportunities and challenges of physics-informed neural networks to serve as a transient stability analysis tool,  highlighting possible pathways to further develop this method.  
    
\end{abstract}

%% file: sections/01_introduction.tex
\section{Introduction}\label{sec:introduction}

Assessing the transient behaviour of power systems is a necessary but difficult process. It belongs to the fundamental operations of power system operators who need to ensure the safe operation of their grid at all times. At frequent intervals, operators assess if probable contingencies result in loss of synchronism, frequency instability, or violations of component limits during the transient phase. The proliferation of converter-connected devices and the continuously increasing penetration of renewable sources, however, add significant complexity to an already complex procedure \cite{hatziargyriou_2021}. Converter-connected devices add substantial non-linearities with time-constants that interfere with the electromagnetic transients of the transmission lines: this requires a much more detailed modeling of the whole grid (EMT simulations) and a much longer computation time to assess each probable contingency. At the same time, renewable sources add significant degrees of uncertainty: this requires the evaluation of a much larger number of possible disturbances related to the active and reactive power balance.


Drastically reducing the computational effort of transient stability simulations while maintaining a high degree of accuracy has been a burning issue and an active research topic for several decades. 
Derived simplified models, e.g., SIME \cite{zhang_sime_1997}, provide more opportunities with respect to screening actions but still, they are often based on linear systems.
Direct methods exploit the state-space characteristics, by using a stability index obtained from energy-like Lyapunov functions \cite{gless_direct_1966, el-abiad_transient_1966}. Thereby, they offer a fast method to assess system stability; however, the resulting certified stable regions are often considered overly conservative, while most approaches, with the exception of \cite{vu_lyapunov_2016}, require linear models.

Machine learning, and neural networks (NN) in particular, might offer relief by providing new approaches and perspectives at these problems as reviewed in \cite{duchesne_recent_2020}. However, these methods suffer from lack of trust on the side of grid operators due to the NN's black-box nature and them being agnostic to the well-studied physical models. In contrast to the conventional methods which primarily derive from the first-principles models that power system engineers have developed for decades, machine learning approaches have been so far based only on data. 
Yet, a high quality dataset can rarely be achieved from data collected from real-world systems. That is because the available real data is usually not enough and, most importantly, it does not represent both normal and abnormal situations equally well, as abnormal events tend to be rare. 
Hence, in order to develop machine learning approaches that could be considered competitive to the conventional methods, one needs to revert to simulations to complement real data. Although the efficiency of generating such databases has been improved \cite{thams_efficient_2020, venzke_datasetcreation_2021}, their high computational burden cannot be disregarded. 

To counter this, recent efforts try to incorporate the physical knowledge collected in models directly into the neural networks. Ref.~\cite{raissi_physics-informed_2018} introduced the governing equations of non-linear systems in the training process of a neural network, referring to them as physics-informed neural networks or PINNs. Our previous work, Ref.~\cite{misyris_physics-informed_2020}, was the first to introduce physics-informed neural networks for power systems. By essentially approximating the solution to ordinary differential equations (ODEs) without the need to perform time-domain simulations, PINNs can drastically reduce the computation time (e.g. by 100x) and allow the screening of more than a hundred contingencies at the same time that conventional methods would only screen a single one. Most importantly, physics-informed neural networks do not rely as much on data. Incorporating the differential-algebraic equations inside the neural network training, PINNs can learn through the incorporated first-principles models instead of using external massive training datasets. 

This paper takes a deep dive into physics-informed neural networks and their application to the transient stability analysis problem. Our goal is to better understand how PINNs perform with respect to computation time, accuracy, and training time compared with both the conventional time-domain simulations and the standard neural networks. For that, we adopt and extend the training procedure to a multi-state system and introduce an additional physcis-based loss. The implementation allows us to determine how different elements of the PINNs contribute to their performance. To maintain a good overview of the inner workings of the different methods and how these behave, we focus our analyses on the Kundur 11-bus system. At the end of the paper we identify the opportunities and challenges of physics-informed neural networks to serve as a transient stability analysis tool, highlighting possible pathways to further develop this method. To further foster the understanding and development of PINNs, we provide a publicly-hosted and open-source code base of the presented work at \url{github.com/jbesty}.

We begin by introducing the methodology in \cref{sec:methodology} and the case study to which we apply the method in \cref{sec:case_study}. In \cref{sec:results}, we extensively describe the inner workings of PINNs before pointing out opportunities and challenges associated with PINNs in \cref{sec:discussion}. \cref{sec:conclusion} concludes.

%% file: sections/02_methodology.tex
\section{Methodology}\label{sec:methodology}

The governing equations of the transient stability problem in power systems are described by \eqref{eq:psdyn1}-\eqref{eq:psdyn2}. Equation \eqref{eq:psdyn1} collects the differential equations, which usually describe the generator and converter dynamics (and possibly the line dynamics if the focus is on the electromagnetic transients), while \eqref{eq:psdyn2} collects the algebraic equations, which usually capture information about the network and the relationship between voltages, currents, and power flows \cite{stott_power_1979}. Vector \xtrue{} represents the dynamic states, which in their simplest form express the rotor angle and frequency of each generator at any time instant, and vector \inputVariable{} describes the system input, e.g. active power setpoints or disturbances.
\begin{align}
    \centering
    \xtruedt &= \vectorf \left(t, \xtrue(t), \inputVariable \right)\label{eq:psdyn1}\\
    \mathbf{0} &= \vectorg \left(t, \xtrue(t), \inputVariable \right) \label{eq:psdyn2}
\end{align}
By specifying a pair $(t_0, \xinitial)$ which represents the initial conditions of the system, 
we can obtain a unique solution for \eqref{eq:psdyn1}-\eqref{eq:psdyn2} along time, that we call a \textit{trajectory}. While the ODE formulation, as shown in \eqref{eq:psdyn1}-\eqref{eq:psdyn2}, yields a `nice' form, a trajectory can not necessarily be represented in a closed form solution. Instead, ODE solvers determine the trajectory by performing the integration numerically over small time steps. As this can be computationally expensive, we aim to find an explicit expression 
\begin{align}
    \xhat(t, \xinitial, \inputVariable) &\approx  \xtrue(t) \begin{cases}  \xtruedt = \vectorf \left(t, \xtrue(t), \inputVariable \right) \\ \xtrue(t_0 = 0) = \xinitial  \end{cases}\label{eq:explicitExpression}
\end{align}
that approximates the trajectory specified by the time instance $t$, the initial condition \xinitial{} and the input variable \inputVariable{} across their input domain; for this purpose we use a neural network (NN).

\subsection{Neural networks as function approximators}

The idea to use NNs as function approximators is well-established. In fact, NNs can theoretically approximate any function \cite{cybenko_approximation_1989, hornik_multilayer_1989}, which sets NN apart from polynomials for example. The practical limitation arises due to the limited representation capacity of the NN which is determined by its size, namely the number of hidden layers \numberLayers{} and the number of neurons per layer \neuronsPerLayer{}, as well as the non-linear activation function \activationFunction{}. We use a plain multi-layer feedforward NN that can be formulated as
\begin{alignat}{2}
    \begin{bmatrix} t, \xinitial, \inputVariable \end{bmatrix}^\top &= \perceptronindex{0} && \label{eq:NN1}\\
    \perceptronindex{k+1} &= \activationFunction \left( \weightindex{k+1} \perceptronindex{k} + \biasindex{k} \right) && \forall k = 0, 1, ..., K-1\label{eq:NN2}\\
    \xhat &= \weightindex{K+1} \perceptronindex{K} + \biasindex{K+1}.\label{eq:NN3}
\end{alignat}
We determine the NN parameters, that is the weights \weights{} and biases \biases{}, in a supervised fashion by minimising the loss \lossxi{}
\begin{align}
    \lossxi &=  \frac{1}{\numberDatapoints}\sum_{j=1}^{\numberDatapoints} (\xtrueij - \xhatij)^2.
\end{align}
We evaluate this loss separately for each dynamic state, indicated by the superscript $i$. Each of the \numberDatapoints{} data points, indexed by the subscript $j$, provides a mapping of the ground truth $(t_j, {\xinitial}_j, \inputVariable_j) \xrightarrow{} \xtrueij$. The training problem then formulates as
\begin{alignat}{2}
    \min_{\weights, \biases} & \quad &&\sum_i \lossWeightxi \lossxi \label{eq:total_loss_NN}\\
    \suchthat & &&\eqref{eq:NN1}-\eqref{eq:NN3}
\end{alignat}
where \lossWeightxi{} provides a weighing of the loss terms (see \cref{subsec:weighing_scheme}).

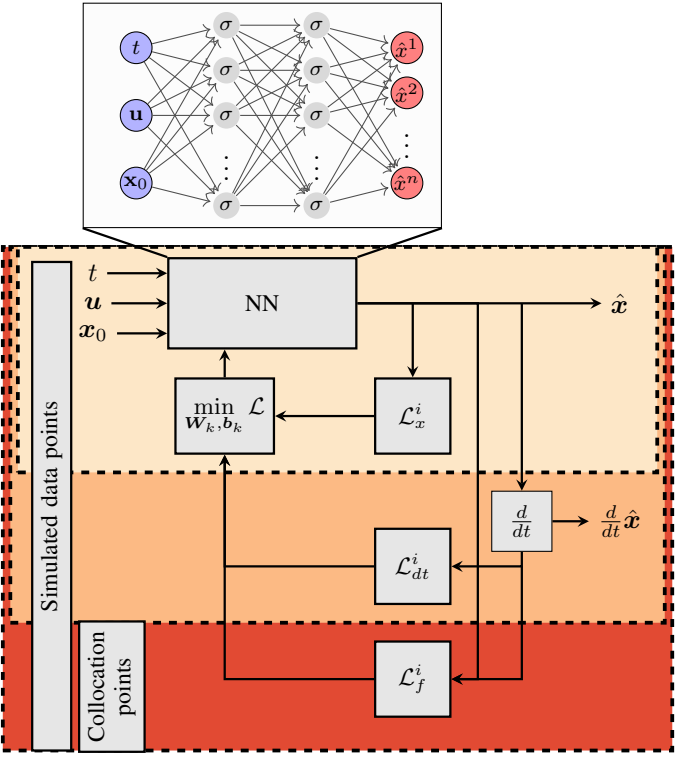
\begin{figure}[!th]
    \centering
    \input{figures/training_process}
    \caption{Structure of the different loss elements in the training procedure of PINNs. The figure shows a simple NN (light orange), the incorporation of the derivative at data points (orange) and the addition of collocation points and the physical loss for PINNs (red).}
    \label{fig:training_process_PINN}
\end{figure}

As we will see in the next subsection, we will keep the structure of the NN \eqref{eq:NN1}-\eqref{eq:NN3} but we update the objective function \eqref{eq:total_loss_NN} to achieve a better prediction accuracy. \Cref{fig:training_process_PINN} illustrates these different steps.

\subsection{Including physics at the data points - dtNN}

In a first step towards using the governing differential equations, we enforce that the state update $\vectorf \left(t, \xtrue(t), \inputVariable \right)$ at the data points matches the temporal derivative of the NN's approximation \xhatdt{}. We calculate \xhatdt{} by applying automatic differentiation (AD) \cite{baydin_automatic_2018} on the NN's outputs \xhat{} with respect to the input time $t$. This yields the loss terms \lossdti{}
\begin{align}
    \lossdti &= \frac{1}{\numberDatapoints}\sum_{j=1}^{\numberDatapoints} \left(\vectorfi(t_j, \xtrue_j,\inputVariable_j) - \xhatdtij \right)^2
\end{align}
which are then weighted and added to \eqref{eq:total_loss_NN}
\begin{alignat}{2}
    \min_{\weights, \biases} & \quad &&\sum_i \lossWeightxi \lossxi + \sum_i \lossWeightdti \lossdti \label{eq:total_loss_dtNN}\\
    \suchthat & &&\eqref{eq:NN1}-\eqref{eq:NN3}.
\end{alignat}
We subsequently refer to this formulation as `dtNN'.

\subsection{Probing physics at collocation points - PINN}
The previous step of introducing dtNNs wrings out more information from the provided data points and their `target values' $\xtrue_j$ than a simple NN. However, the quality of the NN's approximation is still purely dependent on the provided data points. PINNs, in contrast, evaluate the NN's approximation itself with regard to how well it matches the governing equations. This works by comparing the state update based on the approximations $\vectorfi(t_j, \xhat_j,\inputVariable_j)$ and the derivative of the approximated state \xhatdtij{}. The difference yields the loss term \lossfi{}
\begin{align}
    \lossfi &= \frac{1}{\numberCollocationpoints}\sum_{j=1}^{\numberCollocationpoints} \left(\vectorfi(t_j, \xhat_j,\inputVariable_j) - \xhatdtij \right)^2.
\end{align}
Since the calculation of \lossfi{} does not require \xtrue{}, we can evaluate \lossfi{} at any point within the input domain, we refer to them as collocation points (\numberCollocationpoints{}). The training process is now updated once more to include \lossfi{} weighted by \lossWeightfi{}
\begin{alignat}{2}
    \min_{\weights, \biases} & \quad &&\sum_i \lossWeightxi \lossxi + \sum_i \lossWeightdti \lossdti + \sum_i \lossWeightfi \lossfi \label{eq:total_loss_PINN}\\
    \suchthat & &&\eqref{eq:NN1}-\eqref{eq:NN3}.
\end{alignat}

The intuition behind \lossfi{} is as follows: We probe whether the prediction of the NN \xhat{} and its derivative \xhatdt{} are in themselves consistent with the governing equations. However, this does not necessarily mean that the prediction \xhat{} is correct if $\lossfi = 0$. If, for example, the PINN outputs a constant value for \xhat{} with respect to time $t$, i.e., \xhatdt{} = 0, that exactly equals an equilibrium point, i.e., $\vectorf(t, \xhat, \inputVariable) = 0$, \lossfi{} would equal 0 even if the desired trajectory is not the one of the equilibrium. Hence, training a PINN only by \lossfi{} will not yield satisfactory results. The lack of uniqueness can only be resolved if \lossx{} is included, though, in the spirit of the Picard-Lindelöf Theorem, a single value is sufficient to guarantee uniqueness provided that $\lossf = 0$ along the entire trajectory. The initial condition \xinitial{} can fulfill this requirement and as we use \xinitial{} as an input rather than obtaining it as a result of a simulation, it is possible with PINNs to create the solution approximation \xhat{} without a single simulation.

\subsection{Loss term balancing scheme}\label{subsec:weighing_scheme}

We require a single objective value for the machine learning optimisation algorithm, hence we need to weigh the multiple loss terms, that arise due to the state-wise losses \lossxi{}, \lossdti{}, \lossfi{}, in the total loss functions \eqref{eq:total_loss_NN}, \eqref{eq:total_loss_dtNN}, \eqref{eq:total_loss_PINN}. The choice of how to weight these terms can strongly affect the training process and the obtained NN's accuracy. The cause originates from the gradients
\begin{align}
    \frac{d \lossxi}{d (\weights, \biases)}, \frac{d \lossdti}{d (\weights, \biases)}, \frac{d \lossfi}{d (\weights, \biases)}
\end{align}
which potentially yield values of very different magnitudes. As these gradients govern the adjustments of the NN's weights \weights{} and biases \biases{}, it is desirable that no single loss term disproportionately affects the adjustments. To avoid such cases, that can lead to unfavourable local minima, we scale each loss term by the corresponding loss term weight \lossWeightxi{}, \lossWeightdti{}, \lossWeightfi{}. The balancing of these terms is difficult and we rely as of now on a heuristic presented in \cref{tab:loss_term_weights_PINN}. The algorithm proposed by \cite{wang_understanding_2020}, is used as an indication how the loss term weights shall be chosen, however, a continuous update using the algorithm led to unsatisfactory results as its interference with the NN training algorithm created very imbalanced solutions. For the heuristic of \cref{tab:loss_term_weights_PINN} we therefore included the following considerations to adjust the outcome of the algorithm:
\begin{itemize}
    \item Equal weighing of like-wise states (generator rotor angles, generator frequency, load angles)
    \item Scaling of \lossWeightxi{}, \lossWeightdti{} according to the inter-quartile ranges across the dataset
    \item Adjusting \lossWeightxi{}, \lossWeightdti{} to account for the use of collocation points
\end{itemize}

\begin{table}[!th]
\renewcommand{\arraystretch}{1.3}
\caption{Loss terms weights $\lambda$ in \eqref{eq:total_loss_NN}, \eqref{eq:total_loss_dtNN}, \eqref{eq:total_loss_PINN} with $k = 2.2 \cdot 10^4 \cdot \numberCollocationpoints{} / \numberDatapoints{}$}
\label{tab:loss_term_weights_PINN}
\centering
\begin{tabular}{c|ccc}
 & $\delta_1, \delta_2, \delta_3, \delta_4$ & $\delta_7, \delta_9$ & $\omega_1, \omega_2, \omega_3, \omega_4$ \\ \hline
$\lambda_x$  & $1.0 \cdot k$   & $1.0\cdot k$    & $2.0\cdot k$    \\
$\lambda_{dt}$ & $0.5\cdot k$    & $0.04\cdot k$   & $0.12\cdot k$   \\
$\lambda_c$  & 1000   & 3.0    & 5.0   
\end{tabular}
\end{table}

%% file: figures/training_process.tex
\tikzstyle{BOXY} = [rectangle, rounded corners = 5, minimum width=10, minimum height=1.2cm,text centered, draw=black, fill=white,line width=0.3mm,font=\footnotesize, align=center]

\def\NNxshift{2.5cm}
\def\NNyshift{0.4cm}
\begin{tikzpicture}
    \node (PINN) [minimum width=8.9cm, minimum height=6.7cm,draw=black, dashed, line width=0.5mm, fill=color_PINN] at (1cm, -2.6cm) {};
    \node (NN_derivative) [minimum width=8.7cm, minimum height=5cm,draw=black, dashed, line width=0.5mm, fill=color_dtNN] at (1cm, -1.75cm) {};
    \node (NN_pure) [minimum width=8.5cm, minimum height=3.0cm,draw=black, dashed, line width=0.5mm, fill=color_NN] at (1cm, -0.75cm) {};
    \node (NN) [minimum width=2.5cm, minimum height=1.2cm,text centered, draw=black, fill=black!10,line width=0.3mm,font=\small, align=center] {NN};
    \node (input_t) [left of = NN, xshift = -0.5*\NNxshift, yshift = \NNyshift] {$t$};
    \node (input_u) [left of = NN, xshift = -0.5*\NNxshift, yshift = 0.0cm] {\inputVariable{}};
    \node (input_x0) [left of = NN, xshift = -0.5*\NNxshift, yshift = -\NNyshift] {\xinitial{}};
    \node (output_x) [right of = NN, xshift = 1.5*\NNxshift, yshift = 0.0cm] {\xhat{}};
    \node (output_dx) [right of = NN, xshift = 1.5*\NNxshift, yshift = -2.9cm] {\xhatdt{}};
    
    \node (NNarchitecture) [rectangle, text centered,fill=black!1, align=center, draw=black, above of = NN, yshift = 1.5cm] {\input{figures/NN_architecture_small}};
    
    \node (AD) [rectangle, minimum width=0.8cm, minimum height=0.8cm, text centered,fill=black!10, align=center, draw=black, left of = output_dx ,xshift = -0.3cm, yshift = 0.0cm] {$\frac{d}{dt}$};
    
    \node (optimize) [minimum width=1.0cm, minimum height=1.0cm, text centered, draw=black, fill=black!10,line width=0.3mm,font=\small, align=center, below of = NN, xshift = -0.5cm, yshift = -0.5cm] {$\min\limits_{\bm{W}_k, \bm{b}_k} \mathcal{L}$};
    
    \node (loss data) [minimum width=1.0cm, minimum height=1.0cm, text centered, draw=black, fill=black!10,line width=0.3mm,font=\small, align=center, right of = optimize, xshift = 1.5cm, yshift = -0.0cm] {\lossxi{}};
    \node (loss data dt) [minimum width=1.0cm, minimum height=1.0cm, text centered, draw=black, fill=black!10,line width=0.3mm,font=\small, align=center, right of = optimize, xshift = 1.5cm, yshift = -2.0cm] {\lossdti{}};
    \node (loss physics collocation) [minimum width=1.0cm, minimum height=1.0cm, text centered, draw=black, fill=black!10,line width=0.3mm,font=\small, align=center, right of = optimize, xshift = 1.5cm, yshift = -3.5cm] {\lossfi{}};
    
    \node (data) [rotate=90, minimum height=0.5cm,minimum width=6.5cm,text centered, draw=black, fill=black!10,line width=0.3mm,font=\small, align=center, left of = NN, yshift = 2.8cm, xshift = -1.7cm] {Simulated data points};
    \node (collocation) [rotate=90, text width= 1.5cm, minimum height=0.5cm,minimum width=1.5cm,text centered, draw=black, fill=black!10,line width=0.3mm,font=\small, align=center, left of = NN, yshift = 2cm, xshift = -4.1cm] {Collocation points};
    
    \draw[black,->,>=stealth,line width=0.3mm] (input_t) -- (input_t -| NN.west);
    \draw[black,->,>=stealth,line width=0.3mm] (input_u) -- (input_u -| NN.west);
    \draw[black,->,>=stealth,line width=0.3mm] (input_x0) -- (input_x0 -| NN.west);
    \draw[black,->,>=stealth,line width=0.3mm] (NN.east) -- (output_x);   
    
    \draw[black,line width=0.3mm] (NN.north east) -- (NNarchitecture.south east); 
    \draw[black,line width=0.3mm] (NN.north west) -- (NNarchitecture.south west); 
    
    \draw[black,->,>=stealth,line width=0.3mm] (loss data.north |- NN.east) -- (loss data.north); 
    \draw[black,->,>=stealth,line width=0.3mm] (AD.north |- NN.east) -- (AD.north); 
    \draw[black,->,>=stealth,line width=0.3mm] (AD.east) -- (output_dx); 
    \draw[black,->,>=stealth,line width=0.3mm] (optimize.north) -- (optimize.north |- NN.south);
    \draw[black,->,>=stealth,line width=0.3mm] (AD.south) -- (AD.south |- loss data dt.east) -- (loss data dt.east);
    \draw[black,->,>=stealth,line width=0.3mm] (AD.south |- loss data dt.east) -- (AD.south |- loss physics collocation.east) -- (loss physics collocation.east);

    \draw[black,->,>=stealth,line width=0.3mm] (loss data.west) -- (optimize.east);    
    \draw[black,->,>=stealth,line width=0.3mm] (loss data dt.west) -- (loss data dt.west -| optimize.south) -- (optimize.south);
    \draw[black,->,>=stealth,line width=0.3mm] (loss physics collocation.west) -- (loss physics collocation.west -| optimize.south) -- (optimize.south);
    
    \draw[black,->,>=stealth,line width=0.3mm] (NN.east) --++ (1.6cm, 0) --++ (0, -5cm) -- (loss physics collocation.east);
    
    
\end{tikzpicture}

%% file: figures/NN_architecture_small.tex
\pagestyle{empty}
\def\layersep{1.2cm}
\def\nodeinlayersep{0.6cm}
\begin{tikzpicture}[
   shorten >=1pt,->,
   draw=black!70,
    node distance=\layersep,
    every pin edge/.style={<-,shorten <=1pt},
    neuron/.style={circle,fill=black!25,minimum size=10pt,inner sep=0pt, font=\footnotesize},
    input neuron/.style={neuron, fill=blue!30, minimum size=12pt,draw=black},
    output neuron/.style={neuron, fill=red!50, minimum size=12pt,draw=black},
    hidden neuron/.style={neuron, fill=gray!30},
    operator neuron/.style={neuron, fill=yellow!50, minimum size=25pt,draw=black},
    summation neuron/.style={neuron, fill=gray!50, minimum size=18pt},
    parameter neuron/.style={neuron, fill=green!30, minimum size=40pt,draw=black},
    annot/.style={text width=4em, text centered}
]%
    \node[input neuron] (I-1) at (0,-1.5*\nodeinlayersep) {$t$};
    \node[input neuron] (I-2) at (0,-3*\nodeinlayersep) {$\mathbf{u}$};%
    \node[input neuron] (I-3) at (0,-4.5*\nodeinlayersep) {$\mathbf{x}_0$};%
    \node[hidden neuron] (H1-1) at (1*\layersep,-1*\nodeinlayersep ) {$\sigma$};
    \node[hidden neuron] (H1-2) at (1*\layersep,-2*\nodeinlayersep ) {$\sigma$};
    \node[hidden neuron] (H1-3) at (1*\layersep,-3*\nodeinlayersep ) {$\sigma$};
    \node (H1-4) at (1*\layersep,-4*\nodeinlayersep ) {$\vdots$};
    \node[hidden neuron] (H1-5) at (1*\layersep,-5*\nodeinlayersep ) {$\sigma$};
    \node[hidden neuron] (H2-1) at (2*\layersep,-1*\nodeinlayersep ) {$\sigma$};
    \node[hidden neuron] (H2-2) at (2*\layersep,-2*\nodeinlayersep ) {$\sigma$};
    \node[hidden neuron] (H2-3) at (2*\layersep,-3*\nodeinlayersep ) {$\sigma$};
    \node (H2-4) at (2*\layersep,-4*\nodeinlayersep ) {$\vdots$};
    \node[hidden neuron] (H2-5) at (2*\layersep,-5*\nodeinlayersep ) {$\sigma$};
    \node[output neuron] (O-1) at (3*\layersep,-1.5*\nodeinlayersep) {$\hat{x}^1$};%
    \node[output neuron] (O-2) at (3*\layersep,-2.5*\nodeinlayersep) {$\hat{x}^2$};%
    \node (O-3) at (3*\layersep,-3.5*\nodeinlayersep) {$\vdots$};%
    \node[output neuron] (O-4) at (3*\layersep,-4.5*\nodeinlayersep) {$\hat{x}^n$};%
    \foreach \source in {1,2,3}
        \foreach \dest in {1,...,3,5} 
            \path (I-\source) edge (H1-\dest);%
    \foreach \source in {1,...,3,5}
        \foreach \dest in {1,...,3,5} 
            \path (H1-\source) edge (H2-\dest);%
    \foreach \source in {1,...,3,5}
        \foreach \dest in {1,2,4} 
            \path (H2-\source) edge (O-\dest);%
\end{tikzpicture}

%% file: sections/03_case_study.tex
\section{Case study - Kundur's two-area system}\label{sec:case_study}

This section briefly presents a case on the Kundur two-area system which is well-known in the context of transient stability analysis. Using this familiar case, we can highlight fundamental principles associated with PINNs and their training process.

\subsection{Test case - network}

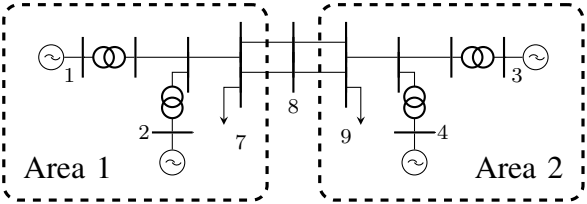
\begin{figure}[htb]
    \centering
    \input{figures/Kundur_2_area_system}
    \caption{Kundur two-area system}
    \label{fig:kundur_2_area}
\end{figure}

\begin{figure}
    \centering
    \input{figures/cascade}
    \caption{Example of a power system response under a load disturbance $\Delta P_7 =\SI{2}{\pu}$ which settled after \SI{5}{\second}. To cover a range of different transient phenomena during our analysis, this is followed by a short circuit at bus 9 lasting for \SI{50}{\milli\second} and the tripping of a line between bus 8 and 9 right afterwards.}
    \label{fig:cascade_trajectory}
\end{figure}
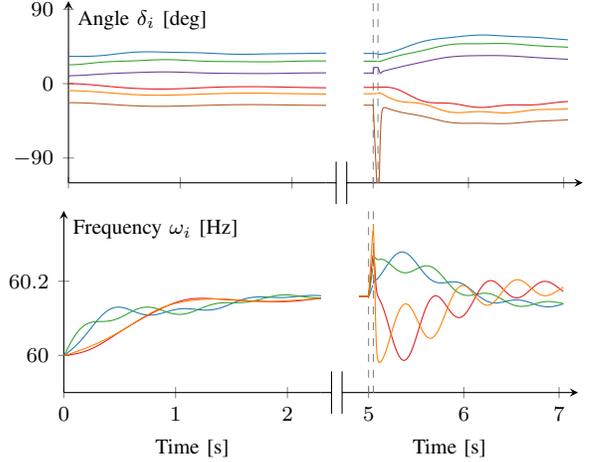

We study a two-area system adopted from Kundur \cite{kundur_power_1994} (refer to page 813 for parameters and a detailed description) as shown in \cref{fig:kundur_2_area} to investigate the interaction between two weakly linked areas under a set of disturbances and faults. We apply the Kron-reduction to eliminate buses without a generator or load connected to it, i.e., purely algebraic equations. The following governing equations then describe the dynamics of buses with a two-order generator model
\begin{align}
    \dot{\delta}_i &= \omega_i - \omega_0 \\
    \dot{\omega}_i &= \frac{\omega_0}{2H_i} \left(P_i^{set}+ \Delta P_i - \sum_j \frac{V_i V_j}{X_{ij}} \sin{(\delta_i - \delta_j)}\right)
    \intertext{and a load respectively}
    \dot{\delta}_i &=  \frac{1}{D_i} \left(P_i^{set} +\Delta P_i - \sum_j \frac{V_i V_j}{X_{ij}} \sin{(\delta_i - \delta_j)} \right)
\end{align}
where the rotor angles $\delta_i$ and the rotor frequencies $\omega_i$ constitute the states $\bm{x}$ of the system. We fix the bus voltages $V_i$ to the values obtained from evaluating an AC-power flow using the fixed power set points $P_i^{set}$. Besides, we are given the inertia and damping constants $H_i$ and $D_i$, as well as the line reactances $X_{ij}$. Lastly, the power disturbance $\Delta P_i$ is the control signal \inputVariable{}, which equals 0 unless stated otherwise. We now apply the following sequence of events: 1) a loss of load at bus 7 $\Delta P_7 \in [0.0, 6.0]\si{\pu}$; 2) after five seconds, when the system is settled: short circuit at bus 9, i.e. $V_9 = \SI{0}{\pu}$. 3) after \SI{50}{\milli\second}: clearing of the short circuit by tripping one line between bus 8 and bus 9, i.e., $X_{8,9}^{new} = 2\cdot X_{8,9}^{old}$. We approximate the evolution after the line trip both with the presented PINNs and compare it to simple NNs and dtNNs. \Cref{fig:cascade_trajectory} illustrates, as an example, the resulting dynamics for a disturbance of $\Delta P_7 =\SI{2}{\pu}$.

\subsection{NN achitecture and training setup}
We use TensorFlow \cite{abadi_tensorflow_nodate} for the implementation of the NNs and the training process utilises the Adam-Optimiser \cite{kingma_adam_2017} with a decaying learning rate\footnote{Please refer to the published code on \url{github.com/jbesty} for details. The initial learning rate is set to 0.025 and the decay leads to reduction between one and two orders of magnitude at the end of the training.}. For all variants of the NNs, we use two hidden layers, 150 nodes per layer, and set the loss term weights as presented in \cref{tab:loss_term_weights_PINN}.
The training data are selected from a simulated database which partitions the input domain [time and power disturbance] into an equally spaced grid with a granularity of \SI{0.001}{\second} across time and \SI{0.002}{\pu} across the size of the power disturbance. The exact training dataset used in the results section will be specified by the number of trajectories $N_{P}$, i.e., each trajectory links to a power disturbance and the number of data points along each trajectory $N_{T}$, hence the total number of data points is $\numberDatapoints = N_{P} \cdot N_{T}$. As far the number of collocation points for PINNs are concerned, these form an equally spaced grid with 25 trajectories and 41 instances along each trajectory. We test all types of NNs on a simulated database that represents a sufficiently fine grid across the input domain, i.e,  $N_{P} = 301$ and $N_{T} = 2001$ corresponding to increments of \SI{0.02}{\pu} with respect to the power disturbance and a temporal resolution of \SI{1}{\milli\second}. Furthermore, we use a validation dataset, which consists of 960 points equally spaced across the input domain, to allow for an early stopping of the training process. All reported timed processes (data creation, training) are performed on a regular machine (i5-7200U CPU @ 2.50GHz, 16GB RAM), whereas the NN training processes for different initialisations are run on a high performance computer of the university.

%% file: figures/Kundur_2_area_system.tex
\pagestyle{empty}
\def\busvdistance{0.7cm}
\def\bushdistancesmall{0.2cm}
\def\bushdistance{1.0cm}

\begin{tikzpicture}[
     vertical bus/.style={rectangle, minimum width=.4pt, minimum height=15pt, inner sep=0pt, draw},
     horizontal bus/.style={rectangle, minimum height=.4pt, minimum width=15pt, inner sep=0pt, draw},
     >=triangle 45,
     every node/.style={font=\footnotesize},
     ]
     
     \tikzset{sin v source/.style={
        circle,
        draw,
        append after command={
            \pgfextra{
                \draw
                ($(\tikzlastnode.center)!0.5!(\tikzlastnode.west)$)
                arc[start angle=150,end angle=30,radius=0.425ex] 
                (\tikzlastnode.center)%
                arc[start angle=210,end angle=330,radius=0.425ex]
                ($(\tikzlastnode.center)!0.5!(\tikzlastnode.east)$) 
                ;
            }
        },
        scale=1.0,
        }
    }
    
    \ctikzset{sources/scale=0.5}

        \node[vertical bus] (bus8) at (0,0) {};
        \node[vertical bus] (bus8s) at (0,-\bushdistancesmall) {};
        \node[vertical bus] (bus8n) at (0,\bushdistancesmall) {};
        \node[vertical bus] (bus7) at (-\busvdistance,0) {};
        \node[vertical bus] (bus7s) at (-\busvdistance,-\bushdistancesmall) {};
        \node[vertical bus] (bus7ss) at (-\busvdistance,-2*\bushdistancesmall) {};
        \node[vertical bus] (bus7n) at (-\busvdistance,\bushdistancesmall) {};
        \node[vertical bus] (bus6) at (-2*\busvdistance,0) {};
        \node[vertical bus] (bus6s) at (-2*\busvdistance,-\bushdistancesmall) {};
        \node[vertical bus] (bus5) at (-3*\busvdistance,0) {};
        \node[vertical bus] (bus1) at (-4*\busvdistance,0) {};
        \node[vertical bus] (bus9) at (\busvdistance,0) {};
        \node[vertical bus] (bus9s) at (\busvdistance,-\bushdistancesmall) {};
        \node[vertical bus] (bus9ss) at (\busvdistance,-2*\bushdistancesmall) {};
        \node[vertical bus] (bus9n) at (\busvdistance,\bushdistancesmall) {};
        \node[vertical bus] (bus10) at (2*\busvdistance,0) {};
        \node[vertical bus] (bus10s) at (2*\busvdistance,-\bushdistancesmall) {};
        \node[vertical bus] (bus11) at (3*\busvdistance,0) {};
        \node[vertical bus] (bus3) at (4*\busvdistance,0) {};
        
        \node[horizontal bus] (bus2) at (-2.3*\busvdistance, -\bushdistance) {};
        \node[horizontal bus] (bus4) at (2.3*\busvdistance, -\bushdistance) {};
        
        \draw (bus8s.west)--(bus7s.east);
        \draw (bus8n.west)--(bus7n.east);
        \draw (bus9s.west)--(bus8s.east);
        \draw (bus9n.west)--(bus8n.east);
        \draw (bus7.west)--(bus6.east);
        \draw (bus6.west)--(bus5.east);
        \draw (bus11.west)--(bus10.east);
        \draw (bus10.west)--(bus9.east);
        
        \draw[->,>=stealth] (bus7ss.west)--++(-0.3*\busvdistance, 0)--++(0,-0.5*\bushdistance); 
        \draw[->,>=stealth] (bus9ss.west)--++(0.3*\busvdistance, 0)--++(0,-0.5*\bushdistance); 
        
        \node[sin v source] (G1) at (-4.6*\busvdistance,0) {};
        \node[sin v source] (G2) at (-2.3*\busvdistance, -1.4*\bushdistance) {};
        \node[sin v source] (G4) at (2.3*\busvdistance, -1.4*\bushdistance) {};
        \node[sin v source] (G3) at (4.6*\busvdistance,0) {};
        \draw (bus1)--(G1.east);
        \draw (bus2)--(G2.north);
        \draw (bus3)--(G3.west);
        \draw (bus4)--(G4.north);
        
        \draw (bus1) to [voosource] (bus5);
        \draw (bus2) to [voosource] (-2.3*\busvdistance,-\bushdistancesmall) -- (bus6s);
        \draw (bus4) to [voosource] (2.3*\busvdistance,-\bushdistancesmall) -- (bus10s);
        \draw (bus3) to [voosource] (bus11);
        
        \node [rounded corners=0.2cm] (area1) at (-3*\busvdistance,-0.6*\bushdistance) [draw,black, dashed, very thick,minimum width=5*\busvdistance,minimum height=2.6*\bushdistance] {};
        \node [rounded corners=0.2cm] (area2) at (3*\busvdistance,-0.6*\bushdistance) [draw,black, dashed, very thick,minimum width=5*\busvdistance,minimum height=2.6*\bushdistance] {};
        \node[anchor=north, yshift = -0.5, xshift=-5.0] at (bus1) {$1$};
        \node[anchor=east, xshift = -5] at (bus2) {$2$};
        \node[anchor=west, xshift = 5] at (bus4) {$4$};
        \node[anchor=north, yshift = -0.5, xshift=5.0] at (bus3) {$3$};
        \node[anchor=north, yshift = -12.5, xshift=0.0] at (bus8) {$8$};
        \node[anchor=north, yshift = -14.5, xshift=0.0] at (bus7ss) {$7$};
        \node[anchor=north, yshift = -12.5, xshift=0.0] at (bus9ss) {$9$};
        
        \node[anchor=north, yshift = -0.6cm, xshift=-0.9cm] at (area1) {\large Area 1};
        \node[anchor=north, yshift = -0.6cm, xshift=0.9cm] at (area2) {\large Area 2};

\end{tikzpicture}

%% file: figures/cascade.tex
\pagestyle{empty}
\def\subplotWidth{6cm}
\def\horizontalDistance{5cm}
\def\subplotHeight{6.5cm}
\def\heightOffset{1.35cm}

\pgfplotsset{scaled x ticks=true}
\begin{tikzpicture}[     every node/.style={font=\footnotesize}]
\begin{groupplot}[
    group style={
    group name=my fancy plots,
    group size=2 by 1,
    xticklabels at=edge bottom,
    horizontal sep=0pt
    },
    ymin=-120, ymax=100,
    ytick={-90, 0, 90},
    height=4cm,
    width=5cm,
    ylabel={Angle $\delta_i$ [deg]},
]

    \nextgroupplot[xmin=0,xmax=2.3,
    xtick={0, 1, 2},
    xticklabels=\empty,
    axis x line=bottom,
    axis y line=middle,
    x axis line style=-,
    ]
    \addplot[draw=color_bus1] table[x index=0, y index=1] {data/cascade_data_disturbance.dat};
    \addplot[draw=color_bus2] table[x index=0, y index=2] {data/cascade_data_disturbance.dat};
    \addplot[draw=color_bus3] table[x index=0, y index=3] {data/cascade_data_disturbance.dat};
    \addplot[draw=color_bus4] table[x index=0, y index=4] {data/cascade_data_disturbance.dat};
    \addplot[draw=color_bus7] table[x index=0, y index=9] {data/cascade_data_disturbance.dat};
    \addplot[draw=color_bus9] table[x index=0, y index=10] {data/cascade_data_disturbance.dat};

    \nextgroupplot[xmin=4.5,xmax=7.2,
    xtick={5, 6, 7},
    xticklabels=\empty,
    axis y line=none,
    axis x line=bottom,
    axis x discontinuity=parallel,
    ]
    
    \draw[color=gray, dashed, very thin] (5,-200) -- (5, 200);
    \draw[color=gray, dashed, very thin] (5.05,-200) -- (5.05, 200);

    \addplot[draw=color_bus1] table[x index=0, y index=1] {data/cascade_data_short_circuit.dat};
    \addplot[draw=color_bus2] table[x index=0, y index=2] {data/cascade_data_short_circuit.dat};
    \addplot[draw=color_bus3] table[x index=0, y index=3] {data/cascade_data_short_circuit.dat};
    \addplot[draw=color_bus4] table[x index=0, y index=4] {data/cascade_data_short_circuit.dat};
    \addplot[draw=color_bus7] table[x index=0, y index=9] {data/cascade_data_short_circuit.dat};
    \addplot[draw=color_bus9] table[x index=0, y index=10] {data/cascade_data_short_circuit.dat};
    
    \addplot[draw=color_bus1] table[x index=0, y index=1] {data/cascade_data_line_trip.dat};
    \addplot[draw=color_bus2] table[x index=0, y index=2] {data/cascade_data_line_trip.dat};
    \addplot[draw=color_bus3] table[x index=0, y index=3] {data/cascade_data_line_trip.dat};
    \addplot[draw=color_bus4] table[x index=0, y index=4] {data/cascade_data_line_trip.dat};
    \addplot[draw=color_bus7] table[x index=0, y index=9] {data/cascade_data_line_trip.dat};
    \addplot[draw=color_bus9] table[x index=0, y index=10] {data/cascade_data_line_trip.dat};
    
    \end{groupplot}

    \end{tikzpicture}
    
\begin{tikzpicture}[     every node/.style={font=\footnotesize}]
    \begin{groupplot}[
    group style={
    group name=my fancy plots,
    group size=2 by 1,
    xticklabels at=edge bottom,
    horizontal sep=0pt
    },
    ymin=59.9, ymax=60.39,
    height=4cm,
    width=5cm,
    xlabel={Time [s]},
    ylabel={Frequency $\omega_i$ [Hz]},
]

    \nextgroupplot[xmin=0,xmax=2.3,
    xtick={0, 1, 2},
    axis x line=bottom,
    axis y line=middle,
    x axis line style=-,
    ]
    \addplot[draw=color_bus1] table[x index=0, y index=5] {data/cascade_data_disturbance.dat};
    \addplot[draw=color_bus2] table[x index=0, y index=6] {data/cascade_data_disturbance.dat};
    \addplot[draw=color_bus3] table[x index=0, y index=7] {data/cascade_data_disturbance.dat};
    \addplot[draw=color_bus4] table[x index=0, y index=8] {data/cascade_data_disturbance.dat};

    \nextgroupplot[xmin=4.5,xmax=7.2,
    xtick={5, 6, 7},
    axis y line=none,
    axis x line=bottom,
    axis x discontinuity=parallel,
    ]
    
    \draw[color=gray, dashed, very thin] (5,59) -- (5, 61);
    \draw[color=gray, dashed, very thin] (5.05,59) -- (5.05, 61);

    \addplot[draw=color_bus1] table[x index=0, y index=5] {data/cascade_data_short_circuit.dat};
    \addplot[draw=color_bus2] table[x index=0, y index=6] {data/cascade_data_short_circuit.dat};
    \addplot[draw=color_bus3] table[x index=0, y index=7] {data/cascade_data_short_circuit.dat};
    \addplot[draw=color_bus4] table[x index=0, y index=8] {data/cascade_data_short_circuit.dat};
    
    \addplot[draw=color_bus1] table[x index=0, y index=5] {data/cascade_data_line_trip.dat};
    \addplot[draw=color_bus2] table[x index=0, y index=6] {data/cascade_data_line_trip.dat};
    \addplot[draw=color_bus3] table[x index=0, y index=7] {data/cascade_data_line_trip.dat};
    \addplot[draw=color_bus4] table[x index=0, y index=8] {data/cascade_data_line_trip.dat};
    
    \end{groupplot}
    \end{tikzpicture}

%% file: sections/04_results.tex
\section{Results}\label{sec:results}

To assess the performance of PINNs, we first compare them with a classical numerical solver with respect to their solution time. In a second step, we show that PINNs are desirable over classical NNs and dtNNs as they achieve a higher accuracy and significantly lower maximum prediction errors for a given training dataset. We proceed by providing further insights into the reasons and the consequences for the training procedure.

\subsection{Evaluation time of PINNs and classical numerical solvers}\label{subsec:PINN_vs_solver}
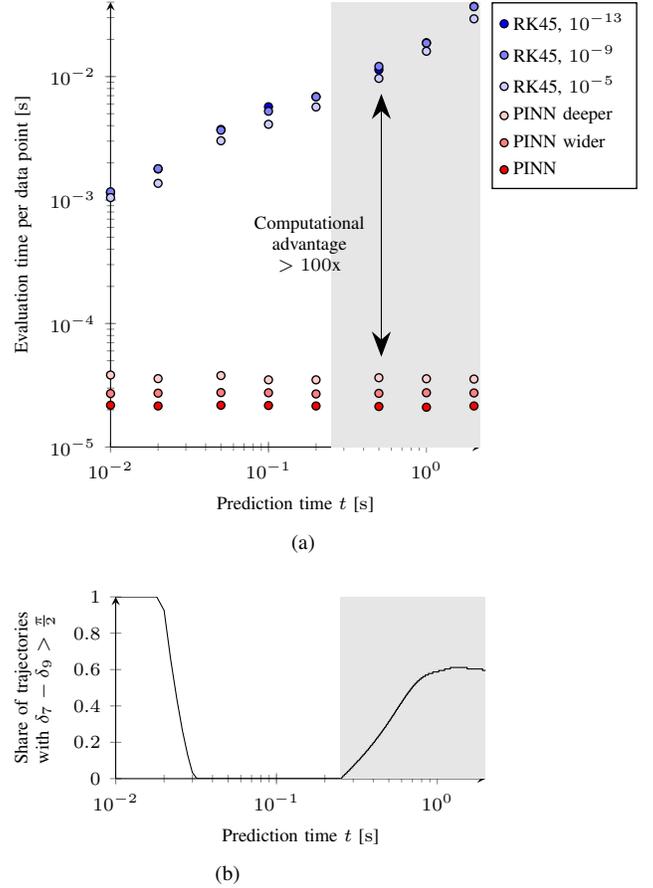
\begin{figure}[!ht]
\subfloat[]{\input{figures/evaluation_time}%
\label{fig:evaluation_time}}\\
\subfloat[]{\input{figures/critical_trajectories}%
\label{fig:critical_trajectories}}
\caption{(a) Comparison of the evaluation time of PINNs (red) and an RK45 solver (blue) with different levels of accuracy and sizes of the NN. (b) critical time to simulate ($>$ \SI{0.25}{\second}) where the first trajectories experience unstable system dynamics.}
\label{fig:PINNs_vs_solver}
\end{figure}

The primary motivation to use forms of NNs, here PINNs, for analysing transient stability behaviour is the speed of evaluation. \Cref{fig:evaluation_time} presents a direct comparison of the evaluation time of PINNs and a classical RK45-solver as implemented in \texttt{scipy.integrate} for various prediction times. While PINNs are about 50 times faster for simulation time periods in the range of \si{\milli\second}, they demonstrate a speed-up of up to 1'000 times faster compared with conventional methods when they need to simulate from $t=\SI{0}{\second}$ to $t=\SI{2}{\second}$. For the present case, the critical value to observe is the angle difference $\delta_7 - \delta_9$, i.e., if the two areas remain synchronised or not. \Cref{fig:critical_trajectories} shows the share of trajectories (in the test dataset) that exhibit $\delta_7 - \delta_9 > \frac{\pi}{2}$, beyond which the system has very high probability to become unstable. Since these critical trajectories are only identifiable after around \SI{0.25}{\second}, even methods such as SIME need to run time-domain simulations for at least up to this region of interest (marked in grey in the figures). Hence, PINNs practically yield a speed up of two to three orders of magnitude in the evaluation for this case, which in turn can allow us to evaluate 100-1'000 more contingency scenarios in the same time frame we would have evaluated only one. 

The reason for this behaviour lies in the fact that PINNs require only a single pass through the network. This links back to the idea to express the output \textit{trajectory} as an explicit expression as in \eqref{eq:explicitExpression} depending on various input values, one of which is the queried time instance. Thereby, the time instance becomes a simple parameter in the evaluation and does not affect the evaluation time. In contrast, the numerical integrator needs successive passes until it reaches the desired time instance. The number of passes depends on an appropriately chosen step size that is dictated by the numerical stability of the solver. A substantial part of the computational expense of the numerical solver lies in the initial steps that are governed by a larger non-linearity and hence require smaller step sizes. From a complexity analysis perspective, the evaluation time of PINNs depends solely on the size of the NN, i.e., the number of layers and the number of neurons per layer. \Cref{fig:evaluation_time} also shows the effects of a double as wide network, i.e. more neurons per layer, and a double as deep network, i.e. more layers. These parameters determine the model complexity which is closely linked to how complex the approximated functions can be.
Determining how `large' or complex a NN needs to be, as a function of the number of states, is non-trivial. In contrast, the computational complexity of ODE solvers links directly to the number of states in the system due to the evaluation of \eqref{eq:psdyn1}. For stiff systems the need for implicit solvers and matrix inversion aggravate this. Furthermore, the tolerance of the ODE solver affects the solution time as shown with the blue dots in \cref{fig:evaluation_time}, whereas for PINNs the level of accuracy is governed by the training process and cannot be varied during evaluation.
The results as presented hold for all three forms of NN, i.e. classical NNs, dtNNs, and PINNs, because the evaluation of \xhat{} as shown in \cref{fig:training_process_PINN} is identical. However, they differ in the training process which governs the accuracy as presented in the subsequent sections.


\subsection{The superior prediction accuracy of PINNs}
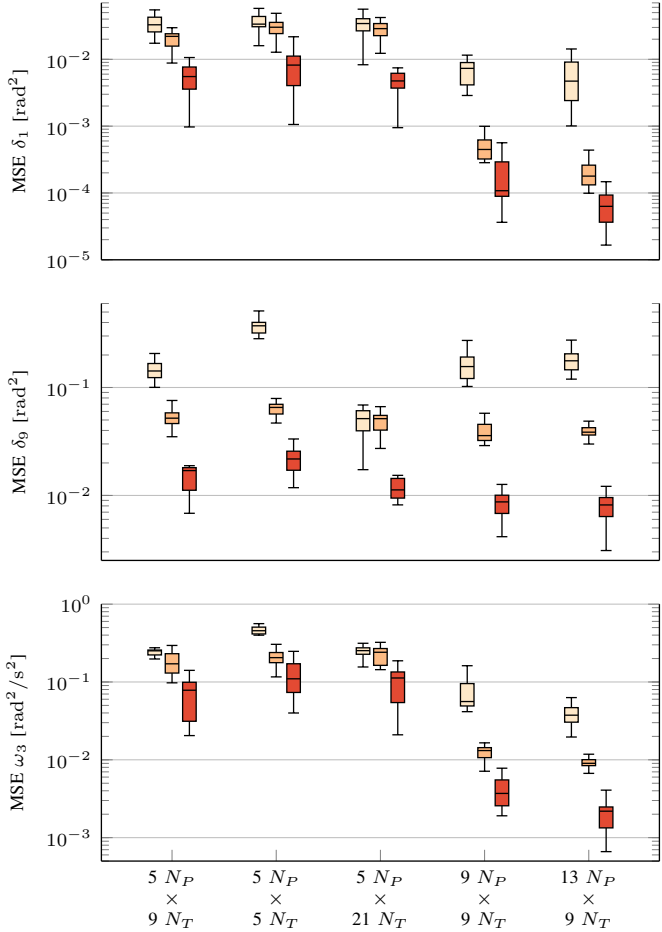
\begin{figure}[!th]
\centering
\input{figures/error_mean}
\caption{Mean squared error (MSE) on the test set for different number of trajectories $N_{P}$ and points per trajectory $N_{T}$, and neural network types (yellow - NN, orange - dtNN, red - PINN). Boxplots represent 20 random initialisations of \weights{} and \biases{}.}
\label{fig:error_boxplots}
\end{figure}

As we have established the advantage of methods based on NNs over classical solvers in terms of evaluation speed, we need to consider how to achieve a required accuracy across the input domain for NN-based methods. Subsequently, the focus lies on the comparison of PINNs with classical NNs and dtNNs with respect to the achievable accuracy and the reasons for the observed behaviour. Therefore, we ask: 'What level of accuracy can we achieve with a given dataset with each method?' \Cref{fig:error_boxplots} provides the answer for a number of dataset sizes. These datasets vary in the number of \textit{trajectories} $N_P$ they contain and the number of data points per \textit{trajectory} $N_T$; in each case the data points are evenly spread across the input domain. The boxplots represent the distribution of the mean squared error for 20 training runs with different, random NN initialisations to show the robustness of the approach.

The main insight of \cref{fig:error_boxplots} is that PINNs consistently outperform simple NNs and dtNNs across all variables and the tested combinations of $N_P \times N_T$. \Cref{fig:error_boxplots} provides us further with two more insights:
1) The comparison of dtNNs and PINNs underlines the fundamental importance of the collocation points and that PINNs are more than simply evaluating physics at `normal' data points. Only the evaluation of \lossfi{} at the collocation points allows to improve \xhat{} for a given limited dataset. 2) Changing the number of trajectories $N_{P}$ or the number of data points per trajectory $N_{T}$ has different effects. For more trajectories, the approximation error reduces, in particular for dtNNs, whereas only using more data points per trajectory leads to insignificant accuracy improvements. The case of $\delta_9$ constitutes an exception to this reasoning, since the dynamics are so fast that a larger $N_T$ is required to capture the dynamics. At this point, PINNs show their full potential since the evaluation of \lossfi{} for $\delta_9$ at the collocation points allows PINNs to better capture these fast dynamics.

The observed effects can be explained when considering the underlying system of differential equations. Points on a given trajectory are governed by temporal derivatives, hence they are linked to each other. Meanwhile, different trajectories are not directly linked to each other apart from the underlying equations which leaves the interpolation between trajectories entirely up to the NN. Due to this fundamental difference of the data points, or more precisely due to their marginal information contribution, using physics on data points improves the accuracy only if it can add information beyond the one provided by the additional data points. In case of more data points per trajectory, the contribution of the data points and the physics overlap and therefore we see no significant improvement in the approximation accuracy. However, for more trajectories using physics provides a clear benefit over a simple NN.

\subsection{Interpolation control}\label{subsec:interpolation_control}
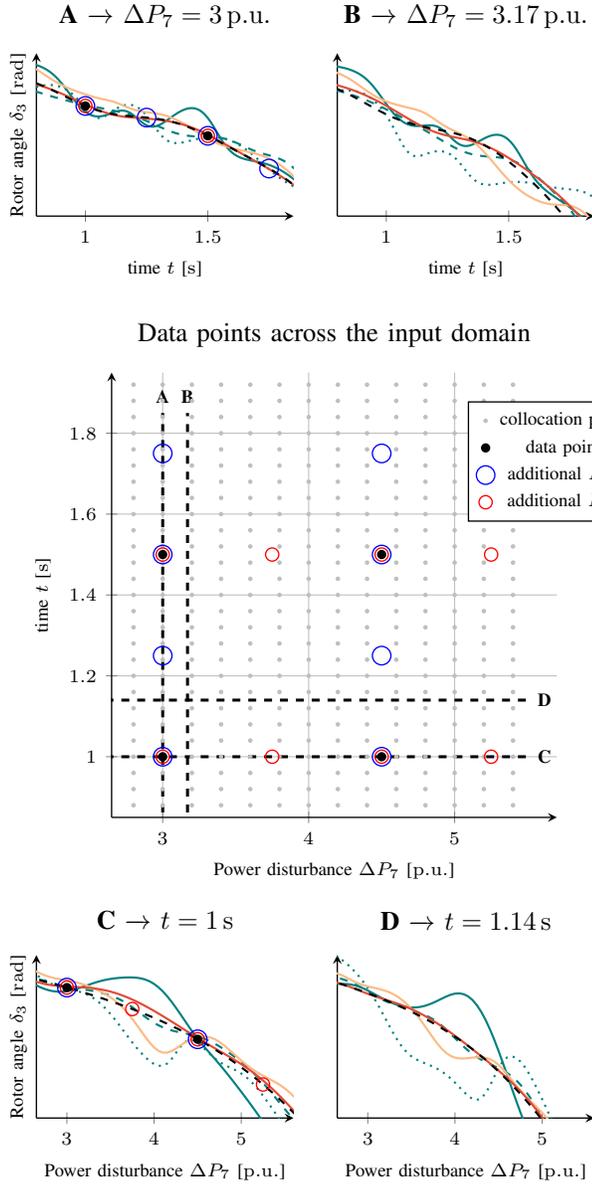
\begin{figure}[!htbp]
\centering
\input{figures/plot_B_data_grid}
\caption{Different number and types of data points and their effect on the interpolation problem across the time and power disturbance domain. Colours: green = simple NN, green dashed = simple NN with more trajectories, green dotted = simple NN with more points per trajectory, orange = dtNN, red = PINN, black dashed = ground truth}
\label{fig:interpolation_NN_type}
\end{figure}

To illustrate the added value of additional data points and collocation points (for PINNs), we consider in \cref{fig:interpolation_NN_type} a section from the two-dimensional input domain (time and power disturbance at bus 7 $\Delta P_7$). The black dots mark the data points that we use for a simple NN and the coloured circles show the used data points in case of more data points per trajectory (red) and more trajectories (blue). Lastly, the grey dots indicate the collocation points used for the PINN together with the black dots. The vertical dashed lines represent `cross-sections' along a trajectory. The two upper plots show these trajectories for $\delta_3$ where the left plot (A) lies on a trajectory with data points while the right one (B) does not, hence is fully subject to the NN's interpolation. We observe see that a NN with more time steps (green dotted line) fits the true trajectory better when looking at the left plots, however, the right plots show a less accurate prediction. Similarly, the dtNN (orange line) shows some mismatches in the right plots. In contrast a NN with more trajectories (green dashed line) and PINNs (red line) provide better approximations in both cases. Similarly, we can consider specific time instances, the horizontal dashed lines in \cref{fig:interpolation_NN_type}. These `snapshots' shown in the plots pronounce the value of more trajectories versus more data points per trajectory even more, as the other NNs apart from PINNs show large errors in their interpolation.

\begin{figure}[!htbp]
\centering
\subfloat[]{\includegraphics[width=0.95\linewidth]{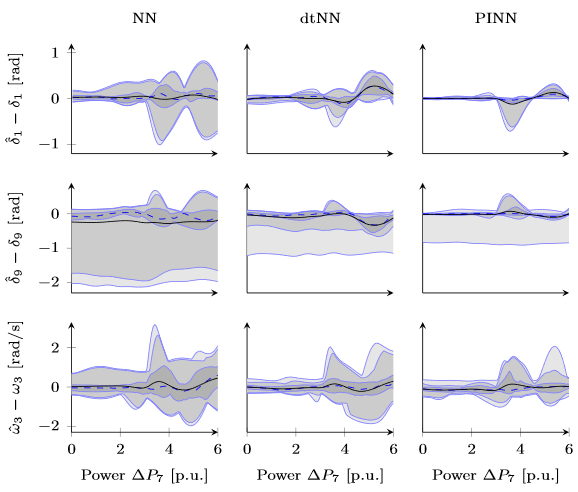}%
\label{fig:power_error}}\\
\subfloat[]{\includegraphics[width=0.95\linewidth]{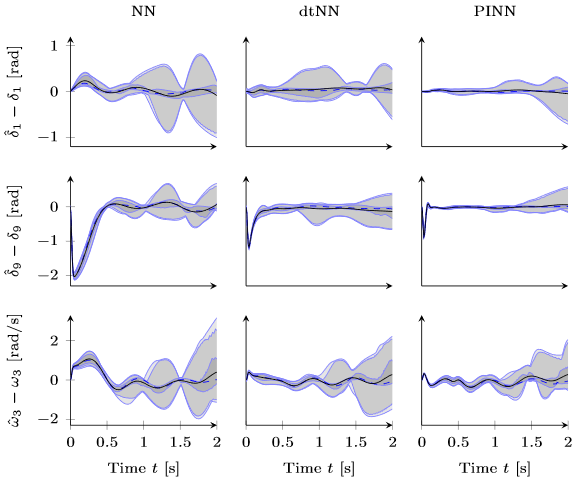}%
\label{fig:time_error}}
\caption{Error distribution of NN, dtNN, PINN for 5 $N_P$, 9 $N_T$ across the power (a) and time input domain (b). The shadings represent the inner 50\%, 90\%, and 100\% of the error distibution, the dashed line shows the median error and the solid black line the mean error.}
\label{fig:error_distribution}
\end{figure}

In practice, the dt loss term but mostly the collocation points are of utmost importance for `controlling' the interpolation of the NN between the regular data points. To illustrate this, we consider the prediction error for 5 trajectories and 9 points per trajectory for three states across the time and power inputs for each of the three NN types in \cref{fig:error_distribution}. The different shades of grey represent the bands in which 100\%, 90\% and 50\% of the prediction errors lie. PINNs stand out by the consistently narrower error bands compared with the dtNNs and NNs. In particular, the errors for small values of $t$ in \cref{fig:time_error} show the ability of PINNs to improve the approximation. In this region, faster dynamics, especially in $\delta_9$, govern the trajectories, hence, traditional NN are unable to capture these, given the low number of data points. Although dtNNs perform better than a simple NN, they cannot mitigate these effects entirely. Another noteworthy observation concerns the error distribution for values of $\Delta P_7 > \SI{3}{\pu}$ in \cref{fig:power_error}. For these values, the system becomes unstable and the wider error bands indicate the worse interpolations. As the function approximation becomes more difficult and complex, PINNs clearly outperform NNs and dtNNs in controlling the error band width.

\subsection{Computation time}\label{subsec:compuation_time}
In this last subsection, we take a look at PINNs with respect to their computational effort. From a holistic perspective, the total computation time to train a NN, dtNN, or PINN consists of three elements: 1) data creation time, 2) training time, 3) testing time. 
The data creation is directly linked to the computational effort associated with using a classical solver as explained in \cref{subsec:PINN_vs_solver}, however, we have to distinguish once more between the number of trajectories $N_P$ versus the number of points per trajectory $N_T$. The number of trajectories $N_P$ primarily drives the data creation time, whereas the number of points per trajectory plays only a minor role. In turn, as we showed earlier (\cref{fig:error_boxplots}), the `cheap' points $N_T$ have also less impact on the accuracy. Given these data points, which we use for NNs, dtNNs, and PINNs, the additional computations that dtNNs and PINNs require in the data creation are negligible, since they only entail a simple evaluation of \vectorf{} for dtNNs and a sampling of the collocation points for PINNs.



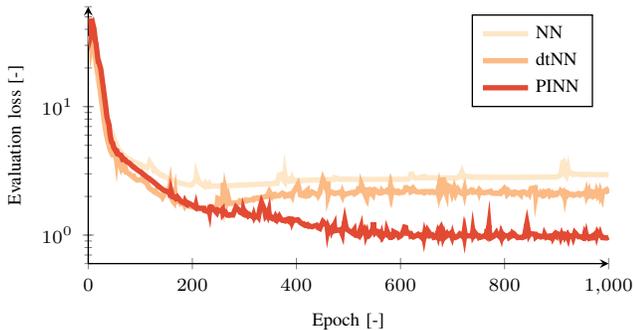
\begin{figure}[!htbp]
    \centering
    \input{figures/validation_loss}
    \caption{Evolution of the validation loss of NN, dtNN, PINN}
    \label{fig:evolution_loss}
\end{figure}

The second component of the overall computation time concerns the training process. The training time is governed by two components: 1) the number of epochs and 2) the time per epoch. \Cref{fig:evolution_loss} shows the evolution of the loss on the validation dataset which serves as a check when the model begins to overfit. Until epoch 200 all three NN types quickly reduce the loss, however, afterwards only the PINNs can improve the performance on the validation set. Hence, continuing the training process leads to improved accuracy at the expense of longer computation time. Due to the additional evaluations of the collocations, each epoch of the PINN is furthermore associated with a four times larger training time per epoch (from about 55ms to 220ms). Ultimately, the benefits of PINNs need to be `paid' in longer training times, however, this trade-off is a matter of choice. In contrast, simple NNs and dtNNs cannot do better without more data points, hence everything hinges around improving the data creation process. 

The third component of the overall computing time is the evaluation and testing of the model. These are crucial to quantify the level of accuracy that can be achieved. The evaluation itself is cheap apart from the necessary test data points as we face the same challenges as in the data creation for the training. This entails that an extensive testing of the model to ensure accurate predictions can be expensive. PINNs do not fundamentally change this problem, however, the higher consistency in the prediction as shown in \cref{subsec:interpolation_control} could be advantageous to reduce the number of required data points. Furthermore, the evaluation of the physics loss \lossfi{}, which does not require the simulation of a trajectory, might be helpful in reducing the computation burden of the testing effort by indicating regions of larger errors. We discuss this opportunity in the following section.





%% file: figures/evaluation_time.tex
\pagestyle{empty}

\def\subplotWidth{6.5cm}
\def\horizontalDistance{5cm}
\def\subplotHeight{7.5cm}
\def\heightOffset{0.5cm}

\begin{tikzpicture}[every node/.style={font=\scriptsize},
    x=0cm,
    y=1cm]%
    \begin{loglogaxis}[
    xlabel near ticks, 
    ylabel near ticks, 
    width=\subplotWidth, 
    height=\subplotHeight,
    yshift=0,
    xshift=0,
    line width=0.5,
    ylabel={Evaluation time per data point [\si{\second}]},
    ymajorticks=true,
    axis x line=bottom,
    axis y line=left,
    xlabel={Prediction time $t$ [\si{\second}]},
    ymin = 0.00001,
    ymax = 0.04,
    xmin = 0.01,
    xmax = 2.2,
    legend style={font=\footnotesize}, 
    legend pos=outer north east,
    legend cell align={left},
    legend columns=1,
    ]%
    \path[name path = leftBoundary] (0.25,0.00001) -- (0.25,0.04);
    \path[name path = rightBoundary] (2.2,0.00001) -- (2.2,0.04);
    \addplot[only marks, mark size=1.5pt, fill=blue] table[x index=0, y index=3,col sep=comma] {data_evaluation_time.dat};
    \addlegendentry{RK45, $10^{-13}$};
    \addplot[only marks, mark size=1.5pt, fill=blue!50] table[x index=0, y index=2,col sep=comma] {data_evaluation_time.dat};
    \addlegendentry{RK45, $10^{-9}$};
    \addplot[only marks, mark size=1.5pt, fill=blue!20] table[x index=0, y index=1,col sep=comma] {data_evaluation_time.dat};
    \addlegendentry{RK45, $10^{-5}$};
    \addplot[only marks, mark size=1.5pt, fill=red!20] table[x index=0, y index=6,col sep=comma] {data_evaluation_time.dat};
    \addlegendentry{PINN deeper};
    \addplot[only marks, mark size=1.5pt, fill=red!50] table[x index=0, y index=5,col sep=comma] {data_evaluation_time.dat};
    \addlegendentry{PINN wider};
    \addplot[only marks, mark size=1.5pt, fill=red] table[x index=0, y index=4,col sep=comma] {data_evaluation_time.dat};
    \addlegendentry{PINN};
    \addplot[gray!20] fill between [of = leftBoundary and rightBoundary];
    \end{loglogaxis}
    \draw[{Stealth[scale=2]}-{Stealth[scale=2]}] (3.6cm, 1.2cm)--++(0, 3.5cm);
    \node[anchor=east, align=center] at (3.5cm, 2.7cm){Computational\\advantage\\$> 100$x};
\end{tikzpicture}
    

%% file: figures/critical_trajectories.tex
\pagestyle{empty}

\def\subplotWidth{6.5cm}
\def\horizontalDistance{5cm}
\def\subplotHeight{4cm}
\def\heightOffset{0.5cm}

\begin{tikzpicture}[every node/.style={font=\scriptsize},
    x=0cm,
    y=1cm]%
    \begin{semilogxaxis}[
    xlabel near ticks, 
    ylabel near ticks, 
    width=\subplotWidth, 
    height=\subplotHeight,
    yshift=0,
    xshift=0cm,
    line width=0.5,
    ylabel={Share of trajectories \\ with $\delta_7 - \delta_9 > \frac{\pi}{2}$},
    ylabel style={align=center},
    ymajorticks=true,
    axis x line=bottom,
    axis y line=left,
    xlabel={Prediction time $t$ [\si{\second}]},
    ymin = 0,
    ymax = 1,
    xmin = 0.01,
    xmax = 2,
    ]%
    \path[name path = leftBoundary] (0.25,0) -- (0.25,1);
    \path[name path = rightBoundary] (2,0) -- (2,1);
    \addplot[gray!20] fill between [of = leftBoundary and rightBoundary];
    \addplot[] table[x index=0, y index=1,col sep=comma] {critical_trajectories.dat};
    \end{semilogxaxis}
\end{tikzpicture}
    

%% file: figures/error_mean.tex
\pagestyle{empty}

\def\subplotWidth{9cm}
\def\horizontalDistance{5cm}
\def\subplotHeight{5cm}
\def\heightOffset{0.5cm}

\begin{tikzpicture}[every node/.style={font=\scriptsize},
    x=0cm,
    yshift=1cm]

    \begin{axis}[boxplot/draw direction=y,
        boxplot/draw position={
                2/6 + floor(\plotnumofactualtype/3)
                  + 1/6*mod(\plotnumofactualtype,3)
            },
            boxplot/box extend=0.13,
    xlabel near ticks, 
    ylabel near ticks, 
    width=\subplotWidth, 
    height=\subplotHeight,
    yshift=\heightOffset,
    xshift=0,
    ymode=log,
    line width=0.5,
    ymajorgrids=true,
    axis x line*=bottom,
    ylabel={MSE $\delta_1$ [\si{\radian^2}]},
    ymin = 0.00001,
    ymax = 0.07,
    xmajorticks=false,
        mark=x,
        mark size=1pt,
    ]

    
    
    \foreach \n in {0,...,14} {
	\addplot+[boxplot, fill, draw=black] table[y index=\n] {data/mean_df_final.dat};
    }
    \end{axis}

    \begin{axis}[boxplot/draw direction=y,
        boxplot/draw position={
                2/6 + floor(\plotnumofactualtype/3)
                  + 1/6*mod(\plotnumofactualtype,3)
            },
            boxplot/box extend=0.13,
    xlabel near ticks, 
    ylabel near ticks, 
    width=\subplotWidth, 
    height=\subplotHeight,
    yshift=\heightOffset- \subplotHeight +1cm,
    xshift=0,
    ymode=log,
    line width=0.5,
    ymajorgrids=true,
    axis x line*=bottom,
    ylabel={MSE $\delta_9$ [\si{\radian^2}]},
    ymin = 0.0025,
    ymax = 0.6,
    xtick={0.5, 1.5, 2.5, 3.5, 4.5},
    xmajorticks=false,
        mark=x,
        mark size=1pt,
    ]

    
    \foreach \n in {18,...,32} {
	\addplot+[boxplot, fill, draw=black] table[y index=\n] {data/mean_df_final.dat};
    }
    \end{axis}

    \begin{axis}[boxplot/draw direction=y,
        boxplot/draw position={
                2/6 + floor(\plotnumofactualtype/3)
                  + 1/6*mod(\plotnumofactualtype,3)
            },
            boxplot/box extend=0.13,
    xlabel near ticks, 
    ylabel near ticks, 
    width=\subplotWidth, 
    height=\subplotHeight,
    yshift=\heightOffset- 2*\subplotHeight +2cm,
    xshift=0,
    ymode=log,
    line width=0.5,
    ymajorgrids=true,
    axis x line*=bottom,
    ylabel={MSE $\omega_3$ [\si{\radian^2\per \second^2}]},
    ymin = 0.0005,
    ymax = 1.0,
    xtick={0.5, 1.5, 2.5, 3.5, 4.5},
    xticklabels={%
        {5 $N_{P}$ \\ $\times$ \\ 9 $N_{T}$},%
        {5 $N_{P}$ \\ $\times$ \\ 5 $N_{T}$},%
        {5 $N_{P}$ \\ $\times$ \\ 21 $N_{T}$},%
        {9 $N_{P}$ \\ $\times$ \\ 9 $N_{T}$},%
        {13 $N_{P}$ \\ $\times$ \\ 9 $N_{T}$},%
    },
    every x tick label/.append style={align=center},
        mark=x,
        mark size=1pt,
    ]
    
    \foreach \n in {36,...,50} {
	\addplot+[boxplot, fill, draw=black] table[y index=\n] {data/mean_df_final.dat};
    }
    \end{axis}
\end{tikzpicture}

%% file: figures/plot_B_data_grid.tex
\pagestyle{empty}

\def\subplotWidth{7.5cm}
\def\horizontalDistance{5cm}
\def\subplotHeight{7.5cm}
\def\heightOffset{0.5cm}

\begin{tikzpicture}[every node/.style={font=\scriptsize},
    simpleNN/.style={draw=teal, line width=0.8},
    NNplusTime/.style={draw=teal, dotted, line width=0.8},
    NNplusPower/.style={draw=teal, dashed, line width=0.8},
    dtNN/.style={draw=color_dtNN, line width=0.8},
    PINN/.style={draw=color_PINN, line width=0.8},
    x=0cm,
    y=1cm]

    \begin{axis}[
    xlabel near ticks, 
    ylabel near ticks, 
    width=\subplotWidth/1.5, 
    height=\subplotHeight/2,
    yshift=\heightOffset+\subplotHeight+0.5cm,
    xshift=-\subplotWidth/3+1.5cm,
    line width=0.5,
    ylabel={Rotor angle $\delta_3$ [\si{\radian}]},
    ymajorticks=false,
    axis x line=bottom,
    axis y line=left,
    xlabel={time $t$ [\si{\second}]},
    title={\normalsize \textbf{A} $\xrightarrow{}$ $\Delta P_7 = \SI{3}{\pu}$},
    ymin = -1.2,
    ymax = -0.4,
    xmin = 0.8,
    xmax = 1.85,
    ]
    
    \addplot[simpleNN] table[x index=0, y index=1] {data/plot_B_power_3.dat};
    \addplot[NNplusTime] table[x index=0, y index=2] {data/plot_B_power_3.dat};
    \addplot[NNplusPower] table[x index=0, y index=3] {data/plot_B_power_3.dat};
    \addplot[dtNN] table[x index=0, y index=4] {data/plot_B_power_3.dat};
    \addplot[PINN] table[x index=0, y index=5] {data/plot_B_power_3.dat};
    \addplot[draw=black, dashed, line width=0.8] table[x index=0, y index=6] {data/plot_B_power_3.dat};
    
    \addplot [only marks, mark size=1.5pt] table {
    1.0 -0.6593975
    1.5 -0.8058731
    };
    \addplot [only marks, mark=o, color=blue, mark size=3.5pt] table {
    1.0 -0.6593975
    1.25 -0.7164390
    1.5 -0.8058731
    1.75 -0.9664949
    };
    \addplot [only marks, mark=o, color=red, mark size=2.5pt] table {
    1.0 -0.6593975
    1.5 -0.8058731
    };
    
    \end{axis}
    
    \begin{axis}[
    xlabel near ticks, 
    ylabel near ticks, 
    width=\subplotWidth/1.5, 
    height=\subplotHeight/2,
    yshift=\heightOffset+\subplotHeight+0.5cm,
    xshift=\subplotWidth/3+0.5cm,
    line width=0.5,
    ymajorticks=false,
    axis x line=bottom,
    axis y line=left,
    xlabel={time $t$ [\si{\second}]},
    title={\normalsize \textbf{B} $\xrightarrow{}$ $\Delta P_7 = \SI{3.17}{\pu}$},
    ymin = -1.2,
    ymax = -0.4,
    xmin = 0.8,
    xmax = 1.85,
    ]
    
        \addplot[simpleNN] table[x index=0, y index=1] {data/plot_B_power_3_2.dat};
    \addplot[NNplusTime] table[x index=0, y index=2] {data/plot_B_power_3_2.dat};
    \addplot[NNplusPower] table[x index=0, y index=3] {data/plot_B_power_3_2.dat};
    \addplot[dtNN] table[x index=0, y index=4] {data/plot_B_power_3_2.dat};
    \addplot[PINN] table[x index=0, y index=5] {data/plot_B_power_3_2.dat};
    \addplot[draw=black, dashed, line width=0.8] table[x index=0, y index=6] {data/plot_B_power_3_2.dat};

    \end{axis}
    
        \begin{axis}[
    xlabel near ticks, 
    ylabel near ticks, 
    width=\subplotWidth/1.5, 
    height=\subplotHeight/2,
    yshift=\heightOffset-4cm,
    xshift=-\subplotWidth/3+1.5cm,
    line width=0.5,
    ymajorticks=false,
    axis x line=bottom,
    axis y line=left,
    ylabel={Rotor angle $\delta_3$ [\si{\radian}]},
    xlabel={Power disturbance $\Delta P_7$ [\si{\pu}]},
    title={\normalsize \textbf{C} $\xrightarrow{}$ $t = \SI{1}{\second}$},
    ymin = -1.7,
    ymax = -0.4,
    xmin = 2.65,
    xmax = 5.6,
    ]
    
    \addplot[simpleNN] table[x index=0, y index=1] {data/plot_B_time_1.dat};
    \addplot[NNplusTime] table[x index=0, y index=2] {data/plot_B_time_1.dat};
    \addplot[NNplusPower] table[x index=0, y index=3] {data/plot_B_time_1.dat};
    \addplot[dtNN] table[x index=0, y index=4] {data/plot_B_time_1.dat};
    \addplot[PINN] table[x index=0, y index=5] {data/plot_B_time_1.dat};
    \addplot[draw=black, dashed, line width=0.8] table[x index=0, y index=6] {data/plot_B_time_1.dat};
    
    \addplot [only marks, mark size=1.5pt] table {
    3.0 -0.6593975
    4.5 -1.0718010
    };
    \addplot [only marks, mark=o, color=blue, mark size=3.5pt] table {
    3.0 -0.6593975
    4.5 -1.0718010
    };
    \addplot [only marks, mark=o, color=red, mark size=2.5pt] table {
    3.0 -0.6593975
    3.75 -0.8307988
    4.5 -1.0718010
    5.25 -1.4325964
    };
    
    \end{axis}
    
    \begin{axis}[
    xlabel near ticks, 
    ylabel near ticks, 
    width=\subplotWidth/1.5, 
    height=\subplotHeight/2,
    yshift=\heightOffset-4cm,
    xshift=\subplotWidth/3+0.5cm,
    line width=0.5,
    ymajorticks=false,
    axis x line=bottom,
    axis y line=left,
    xlabel={Power disturbance $\Delta P_7$ [\si{\pu}]},
    title={\normalsize \textbf{D} $\xrightarrow{}$ $t = \SI{1.14}{\second}$},
    ymin = -1.7,
    ymax = -0.4,
    xmin = 2.65,
    xmax = 5.6,
    ]
    
    \addplot[simpleNN] table[x index=0, y index=1] {data/plot_B_time_1_14.dat};
    \addplot[NNplusTime] table[x index=0, y index=2] {data/plot_B_time_1_14.dat};
    \addplot[NNplusPower] table[x index=0, y index=3] {data/plot_B_time_1_14.dat};
    \addplot[dtNN] table[x index=0, y index=4] {data/plot_B_time_1_14.dat};
    \addplot[PINN] table[x index=0, y index=5] {data/plot_B_time_1_14.dat};
    \addplot[draw=black, dashed, line width=0.8] table[x index=0, y index=6] {data/plot_B_time_1_14.dat};
    
    \end{axis}

    

    \begin{axis}[
    xlabel near ticks, 
    ylabel near ticks, 
    width=\subplotWidth, 
    height=\subplotHeight,
    yshift=\heightOffset,
    xshift=0,
    line width=0.5,
    grid=major, 
    ylabel={time $t$ [\si{\second}]},
    xlabel={Power disturbance $\Delta P_7$ [\si{\pu}]},
    title={\normalsize Data points across the input domain},
    ymin = 0.85,
    ymax = 1.95,
    xmin = 2.65,
    xmax = 5.7,
    axis x line=bottom,
    axis y line=left,
    legend style={at={(0.8,0.8)}, anchor=west,legend columns=1},
          legend style={font=\small},
          legend entries={collocation point,
                          data point,
                          additional $N_T$,
                        additional $N_P$,
                          }
    ]

    \addplot[only marks, mark size=0.6pt, gray!50] table {
2.6 0.84
2.6 0.88
2.6 0.92
2.6 0.96
2.6 1.00
2.6 1.04
2.6 1.08
2.6 1.12
2.6 1.16
2.6 1.20
2.6 1.24
2.6 1.28
2.6 1.32
2.6 1.36
2.6 1.40
2.6 1.44
2.6 1.48
2.6 1.52
2.6 1.56
2.6 1.60
2.6 1.64
2.6 1.68
2.6 1.72
2.6 1.76
2.6 1.80
2.6 1.84
2.6 1.88
2.6 1.92
2.6 1.96
2.6 2.00
2.8 0.84
2.8 0.88
2.8 0.92
2.8 0.96
2.8 1.00
2.8 1.04
2.8 1.08
2.8 1.12
2.8 1.16
2.8 1.20
2.8 1.24
2.8 1.28
2.8 1.32
2.8 1.36
2.8 1.40
2.8 1.44
2.8 1.48
2.8 1.52
2.8 1.56
2.8 1.60
2.8 1.64
2.8 1.68
2.8 1.72
2.8 1.76
2.8 1.80
2.8 1.84
2.8 1.88
2.8 1.92
2.8 1.96
2.8 2.00
3.0 0.84
3.0 0.88
3.0 0.92
3.0 0.96
3.0 1.00
3.0 1.04
3.0 1.08
3.0 1.12
3.0 1.16
3.0 1.20
3.0 1.24
3.0 1.28
3.0 1.32
3.0 1.36
3.0 1.40
3.0 1.44
3.0 1.48
3.0 1.52
3.0 1.56
3.0 1.60
3.0 1.64
3.0 1.68
3.0 1.72
3.0 1.76
3.0 1.80
3.0 1.84
3.0 1.88
3.0 1.92
3.0 1.96
3.0 2.00
3.2 0.84
3.2 0.88
3.2 0.92
3.2 0.96
3.2 1.00
3.2 1.04
3.2 1.08
3.2 1.12
3.2 1.16
3.2 1.20
3.2 1.24
3.2 1.28
3.2 1.32
3.2 1.36
3.2 1.40
3.2 1.44
3.2 1.48
3.2 1.52
3.2 1.56
3.2 1.60
3.2 1.64
3.2 1.68
3.2 1.72
3.2 1.76
3.2 1.80
3.2 1.84
3.2 1.88
3.2 1.92
3.2 1.96
3.2 2.00
3.4 0.84
3.4 0.88
3.4 0.92
3.4 0.96
3.4 1.00
3.4 1.04
3.4 1.08
3.4 1.12
3.4 1.16
3.4 1.20
3.4 1.24
3.4 1.28
3.4 1.32
3.4 1.36
3.4 1.40
3.4 1.44
3.4 1.48
3.4 1.52
3.4 1.56
3.4 1.60
3.4 1.64
3.4 1.68
3.4 1.72
3.4 1.76
3.4 1.80
3.4 1.84
3.4 1.88
3.4 1.92
3.4 1.96
3.4 2.00
3.6 0.84
3.6 0.88
3.6 0.92
3.6 0.96
3.6 1.00
3.6 1.04
3.6 1.08
3.6 1.12
3.6 1.16
3.6 1.20
3.6 1.24
3.6 1.28
3.6 1.32
3.6 1.36
3.6 1.40
3.6 1.44
3.6 1.48
3.6 1.52
3.6 1.56
3.6 1.60
3.6 1.64
3.6 1.68
3.6 1.72
3.6 1.76
3.6 1.80
3.6 1.84
3.6 1.88
3.6 1.92
3.6 1.96
3.6 2.00
3.8 0.84
3.8 0.88
3.8 0.92
3.8 0.96
3.8 1.00
3.8 1.04
3.8 1.08
3.8 1.12
3.8 1.16
3.8 1.20
3.8 1.24
3.8 1.28
3.8 1.32
3.8 1.36
3.8 1.40
3.8 1.44
3.8 1.48
3.8 1.52
3.8 1.56
3.8 1.60
3.8 1.64
3.8 1.68
3.8 1.72
3.8 1.76
3.8 1.80
3.8 1.84
3.8 1.88
3.8 1.92
3.8 1.96
3.8 2.00
4.0 0.84
4.0 0.88
4.0 0.92
4.0 0.96
4.0 1.00
4.0 1.04
4.0 1.08
4.0 1.12
4.0 1.16
4.0 1.20
4.0 1.24
4.0 1.28
4.0 1.32
4.0 1.36
4.0 1.40
4.0 1.44
4.0 1.48
4.0 1.52
4.0 1.56
4.0 1.60
4.0 1.64
4.0 1.68
4.0 1.72
4.0 1.76
4.0 1.80
4.0 1.84
4.0 1.88
4.0 1.92
4.0 1.96
4.0 2.00
4.2 0.84
4.2 0.88
4.2 0.92
4.2 0.96
4.2 1.00
4.2 1.04
4.2 1.08
4.2 1.12
4.2 1.16
4.2 1.20
4.2 1.24
4.2 1.28
4.2 1.32
4.2 1.36
4.2 1.40
4.2 1.44
4.2 1.48
4.2 1.52
4.2 1.56
4.2 1.60
4.2 1.64
4.2 1.68
4.2 1.72
4.2 1.76
4.2 1.80
4.2 1.84
4.2 1.88
4.2 1.92
4.2 1.96
4.2 2.00
4.4 0.84
4.4 0.88
4.4 0.92
4.4 0.96
4.4 1.00
4.4 1.04
4.4 1.08
4.4 1.12
4.4 1.16
4.4 1.20
4.4 1.24
4.4 1.28
4.4 1.32
4.4 1.36
4.4 1.40
4.4 1.44
4.4 1.48
4.4 1.52
4.4 1.56
4.4 1.60
4.4 1.64
4.4 1.68
4.4 1.72
4.4 1.76
4.4 1.80
4.4 1.84
4.4 1.88
4.4 1.92
4.4 1.96
4.4 2.00
4.6 0.84
4.6 0.88
4.6 0.92
4.6 0.96
4.6 1.00
4.6 1.04
4.6 1.08
4.6 1.12
4.6 1.16
4.6 1.20
4.6 1.24
4.6 1.28
4.6 1.32
4.6 1.36
4.6 1.40
4.6 1.44
4.6 1.48
4.6 1.52
4.6 1.56
4.6 1.60
4.6 1.64
4.6 1.68
4.6 1.72
4.6 1.76
4.6 1.80
4.6 1.84
4.6 1.88
4.6 1.92
4.6 1.96
4.6 2.00
4.8 0.84
4.8 0.88
4.8 0.92
4.8 0.96
4.8 1.00
4.8 1.04
4.8 1.08
4.8 1.12
4.8 1.16
4.8 1.20
4.8 1.24
4.8 1.28
4.8 1.32
4.8 1.36
4.8 1.40
4.8 1.44
4.8 1.48
4.8 1.52
4.8 1.56
4.8 1.60
4.8 1.64
4.8 1.68
4.8 1.72
4.8 1.76
4.8 1.80
4.8 1.84
4.8 1.88
4.8 1.92
4.8 1.96
4.8 2.00
5.0 0.84
5.0 0.88
5.0 0.92
5.0 0.96
5.0 1.00
5.0 1.04
5.0 1.08
5.0 1.12
5.0 1.16
5.0 1.20
5.0 1.24
5.0 1.28
5.0 1.32
5.0 1.36
5.0 1.40
5.0 1.44
5.0 1.48
5.0 1.52
5.0 1.56
5.0 1.60
5.0 1.64
5.0 1.68
5.0 1.72
5.0 1.76
5.0 1.80
5.0 1.84
5.0 1.88
5.0 1.92
5.0 1.96
5.0 2.00
5.2 0.84
5.2 0.88
5.2 0.92
5.2 0.96
5.2 1.00
5.2 1.04
5.2 1.08
5.2 1.12
5.2 1.16
5.2 1.20
5.2 1.24
5.2 1.28
5.2 1.32
5.2 1.36
5.2 1.40
5.2 1.44
5.2 1.48
5.2 1.52
5.2 1.56
5.2 1.60
5.2 1.64
5.2 1.68
5.2 1.72
5.2 1.76
5.2 1.80
5.2 1.84
5.2 1.88
5.2 1.92
5.2 1.96
5.2 2.00
5.4 0.84
5.4 0.88
5.4 0.92
5.4 0.96
5.4 1.00
5.4 1.04
5.4 1.08
5.4 1.12
5.4 1.16
5.4 1.20
5.4 1.24
5.4 1.28
5.4 1.32
5.4 1.36
5.4 1.40
5.4 1.44
5.4 1.48
5.4 1.52
5.4 1.56
5.4 1.60
5.4 1.64
5.4 1.68
5.4 1.72
5.4 1.76
5.4 1.80
5.4 1.84
5.4 1.88
5.4 1.92
5.4 1.96
5.4 2.00
    };
    \draw[dashed,line width=0.4mm] (3, 0) -- (3,1.85) node[above] {\textbf{A}};
    \draw[dashed,line width=0.4mm] (3.17, 0) -- (3.17,1.85) node[above] {\textbf{B}};
    \draw[dashed,line width=0.4mm] (0, 1) -- (5.5,1)  node[right] {\textbf{C}};
    \draw[dashed,line width=0.4mm] (0, 1.14) -- (5.5,1.14) node[right] {\textbf{D}};
    \addplot [only marks, mark size=1.5pt] table {
    3.0 1.0
    3.0 1.5
    4.5 1.0
    4.5 1.5
    };
    \addplot [only marks, mark=o, color=blue, mark size=3.5pt] table {
    3.0 1.0
    3.0 1.25
    3.0 1.5
    3.0 1.75
    4.5 1.0
    4.5 1.25
    4.5 1.5
    4.5 1.75
    };
    \addplot [only marks, mark=o, color=red, mark size=2.5pt] table {
    3.0 1.0
    3.0 1.5
    3.75 1.0
    3.75 1.5
    4.5 1.0
    4.5 1.5
    5.25 1.0
    5.25 1.5
    };
    \end{axis}
    
\end{tikzpicture}
    

%% file: figures/validation_loss.tex
\pagestyle{empty}

\def\subplotWidth{8.5cm}
\def\horizontalDistance{5cm}
\def\subplotHeight{5cm}
\def\heightOffset{0.5cm}

\begin{tikzpicture}[every node/.style={font=\scriptsize},
    x=0cm,
    y=1cm]%
    \begin{semilogyaxis}[
    xlabel near ticks, 
    ylabel near ticks, 
    width=\subplotWidth, 
    height=\subplotHeight,
    yshift=0,
    xshift=0,
    line width=0.5,
    ylabel={Evaluation loss [-]},
    ymajorticks=true,
    axis x line=bottom,
    axis y line=left,
    xlabel={Epoch [-]},
    ymin = 0.6,
    ymax = 60,
    xmin = 0,
    xmax = 1000,
    legend style={font=\footnotesize}, 
    legend pos=north east,
    legend cell align={left},
    legend columns=1,
    ]%
    \addplot[draw=color_NN, line width=2] table[x index=0, y index=1] {evaluation_loss_data.dat};
    \addlegendentry{NN};
    \addplot[draw=color_dtNN, line width=2] table[x index=0, y index=2] {evaluation_loss_data.dat};
    \addlegendentry{dtNN};
    \addplot[draw=color_PINN, line width=2] table[x index=0, y index=3] {evaluation_loss_data.dat};
    \addlegendentry{PINN};
    \end{semilogyaxis}
\end{tikzpicture}

%% file: sections/05_discussion.tex
\section{Opportunities and challenges for PINNs}\label{sec:discussion}
In the previous sections, we have been investigating a very specific case in depth; let us now take a broader view on this method, while also outline directions for future work.

\subsubsection{Scalability}
The most pressing question is indisputably: Does this method scale? This aspect needs to be considered from two angles: First, increasing the number of buses and hence states, and second, increasing the input dimensionality, i.e., the number of inputs, and their input range. The latter is the harder problem due to the much-cited `curse of dimensionality' and a crucial step will be to select the most influential input variables and limit them to sensible ranges, for which power system `expert knowledge' is essential. Applying PINNs to larger systems but with a low input dimensionality seems much more feasible, based on the following idea: The solution to the ODEs will likely lie on a lower-dimensional manifold which PINNs can exploit. Take the center of inertia of the two areas in the presented Kundur system; this gives already a good approximation of the frequency response of the generators and could be seen as such a manifold. In either case, appropriate machine learning algorithms and an improved understanding of the loss term balancing will be key factors to succeed. The upside, though, also increases, as the computational advantage in the evaluation of the ODEs only will become larger with increasing system sizes.

\subsubsection{Holistic training process and error estimation}
\Cref{subsec:compuation_time} already hinted that PINN offer the opportunity to look differently at the entire training process. Instead of being solely focused on improved data creation methods, PINNs allows us to tie the training and testing closer to the data creation. Such approaches could use the evaluation of \lossfi{} to infer the true approximation errors. In particular, data efficient sampling processes that identify regions in the input domain of higher interest or importance could lead to significant improvements of the entire process. Furthermore, such analyses could lead to providing metrics on the NN's accuracy without the need for extensive simulated test sets. 

\subsubsection{Transfer learning}
Borrowing an idea from the machine learning community, transfer learning could be a direction for PINNs. It means that a learning task starts from a model that was trained on a (slightly) different task beforehand. What if we trained a PINN on an intact system under various load disturbances and then change topology by simply adapting the physical equations. This varied learning task could potentially profit from the previous network's training and hence yield accurate predictions more quickly than when trained from scratch.

%% file: sections/06_conclusion.tex
\section{Conclusion}\label{sec:conclusion}

Machine learning approaches offer enormous opportunities for screening dynamic responses of power systems, thanks to their extremely fast system state evaluations compared to established ODE solvers. As the evaluation time of PINNs, and NNs in general, is decoupled from the accuracy of the approximation, the challenge of applying NNs lies in achieving sufficiently high accuracy. In this work we show how the usage of the governing physical equations fundamentally changes the workflow of applying NNs. Not only do we achieve a more data-efficient training process, PINNs also offer the upside of providing easily computable metrics to identify areas of inaccuracy based on the agreement of the prediction with the physical equations.